\documentclass[preprint, review, 3p, authoryear, times]{elsarticle}

\usepackage{todonotes}
\usepackage{subcaption} 

\usepackage{amsmath}  
\usepackage{capt-of}  

\usepackage{graphicx}
\usepackage{url}
\usepackage{color, soul}
\usepackage{bm}
\usepackage{amsmath} 
\usepackage{lineno}
\usepackage{float}  

\usepackage{amssymb}
 
\usepackage{hyperref}
\usepackage{float}
 \usepackage{graphicx}
\usepackage{graphicx}
\usepackage{longtable}
\usepackage{booktabs}
\usepackage{float}
\usepackage[font=small]{caption}  

\usepackage{placeins}  
\usepackage{xcolor}
\newcommand{\rev}[1]{\textcolor{black}{#1}}

\presetkeys{todonotes}{inline,backgroundcolor=lightblue}{}

\begin{document}

\title{
Generative AI as a Pillar for Predicting 2D and 3D Wildfire Spread: Beyond Physics-Based Models and Traditional Deep Learning
}


\author[UNSW-BE]{Haowen Xu } \ead{haowen.xu1@unsw.edu.au}
\author[UNSW-BE]{Sisi Zlatanova\corref{cor1}}\ead{s.zlatanova@unsw.edu.au}
\author[UNSW-M]{Ruiyu Liang}\ead{ruiyu.liang@unsw.edu.au}
\author[UNSW-M]{Ismet Canbulat} \ead{i.canbulat@unsw.edu.au}

\cortext[cor1]{Corresponding author.}

 
\address[UNSW-BE]{GRID, School of Built Environment, UNSW Sydney, NSW 2052 Australia}
\address[UNSW-M]{School of Minerals and Energy Resources Engineering, UNSW Sydney, NSW 2052 Australia}

\makeatletter
\newcommand{\printfnsymbol}[1]{%
  \textsuperscript{\@fnsymbol{#1}}%
}

\newcommand*{\MyIndent}{\hspace*{0.5cm}}%

\begin{abstract}
Wildfires increasingly threaten human life, ecosystems, and infrastructure, with events like the \rev{2025 Palisades and Eaton fires in Los Angeles County underscoring} the urgent need for more advanced prediction frameworks. Existing physics-based and deep learning models struggle to capture dynamic wildfire spread across both 2D and 3D domains, especially when incorporating real-time, multimodal geospatial data. This paper explores how generative Artificial Intelligence (AI) models—such as GANs, VAEs, and Transformers—can serve as transformative tools for wildfire prediction and simulation. These models offer superior capabilities in managing uncertainty, integrating multimodal inputs, and generating realistic, scalable wildfire scenarios. We introduce a new paradigm that leverages large language models (LLMs) for literature synthesis, classification, and knowledge extraction, \rev{conducting a systematic review of recent studies applying generative AI to fire prediction and monitoring. We highlight how generative approaches uniquely address challenges faced by traditional simulation and deep learning methods.} Finally, we outline five key future directions for generative AI in wildfire management, including unified multimodal modeling of 2D and 3D dynamics, agentic AI systems and chatbots for decision intelligence, and real-time scenario generation on mobile devices, along with a discussion of critical challenges. Our findings advocate for a paradigm shift toward multimodal generative frameworks to support proactive, data-informed wildfire response.
\end{abstract}




\maketitle

\section{Introduction}
Wildfires and bushfires have emerged as one of the most destructive natural disasters of the 21st century, leaving a devastating trail across natural ecosystems, agricultural lands, and densely populated urban regions \cite{bowman2020vegetation, doerr2016global, kotozaki2025increase}. These fast-moving fires not only cause immense structural damage and loss of human life but also result in widespread economic disruption and long-term environmental degradation \rev{\cite{goralnick2025long, amiri2025firestorm, nasa2025fires}}. The release of smoke, fine particulate matter, and toxic gases during large-scale bushfires contributes significantly to air pollution, impacting public health and exacerbating climate change in cities across the globe \cite{johnston2012estimated, urbanski2014wildland}. 
\rev{Striking examples} of this escalating crisis include the 2025 Los Angeles wildfire season—when the Palisades and Eaton fires swept through Southern California—\rev{and Australia’s 2019–2020 “Black Summer” fires} \rev{\cite{brown2025confluence, han2025spatial, woolcott2025angeles, BushfireRecovery2020}}.
Together, they caused an estimated \$250 billion in economic losses, destroyed thousands of homes, and displaced entire communities, inflicting deep physical and emotional suffering on affected populations \citep{EPA2025,HarvardGazette2025}. These events underscore the urgent need for accurate and timely wildfire forecasting systems. In particular, real-time bushfire propagation simulation-capable of predicting fire spread pathways and identifying at-risk zones-plays a critical role in supporting firefighting operations, emergency evacuation planning, and long-term fire hazard mitigation strategies \cite{baptiste2009coupled, barton2024near, wang2019safe, wang2014data}.

A variety of wildfire propagation models have been developed, each rooted in distinct computational approaches. Physics-based models (e.g., FARSITE, SPARK, Prometheus) simulate fire spread using physical laws of combustion, heat transfer, fuel conditions, and wind dynamics \cite{finney1998farsite, mell2007physics, miller2015spark, tymstra2010development, barton2020voxel}, offering detailed outputs but requiring extensive computation and environmental inputs—often limiting their use in real-time scenarios \cite{dipierro2024simple}. Empirical models, like McArthur’s Fire Danger Index, rely on historical fire behavior to produce rapid, region-specific predictions, but struggle with generalizability across diverse ecosystems \cite{noble1980mcarthur}. Traditional machine learning models (e.g., decision trees, support vector machines) have been used for fire ignition and risk classification \cite{jain2020review, bot2022systematic, nur2023spatial, ismail2024assessment}, though their capacity to simulate dynamic fire progression is limited. 

In contrast, deep learning models have recently gained momentum for 2D fire spread forecasting, due to their ability to capture complex spatiotemporal relationships in fire behavior \cite{andrianarivony2024machine, shadrin2024wildfire}. Architectures such as Convolutional Neural Networks (CNNs), Recurrent Neural Networks (RNNs), and U-Net variants have been applied to wildfire segmentation and short-term spread prediction, leveraging remote sensing, meteorological, and topographic data \cite{ghali2023deep, vargas2025development, zakari2025enhanced, alizadeh2024fusionfirenet}. These deep learning models (e.g., CNNs and RNNs) often outperform traditional physics-based simulators in runtime, scalability, and—in data-rich environments—predictive accuracy, as they learn complex patterns and variability directly from empirical observations rather than relying on explicit physical formulations \cite{radke2019firecast, chatterjee2024prescribed}. However, deep learning-based fire models still face key limitations that prevent them from enabling more advanced fire prediction capabilities with higher spatial and temporal resolution in real time. Most models remain constrained to \textbf{2D spatial domains} \cite{andrianarivony2024machine} and \textbf{lack integration and real-time augmentation of multimodal inputs}, such as meteorological, topographic, and vegetation (fuel) data \cite{radke2019firecast, abdollahi2025challenges}, which often change dynamically as fire propagates across the landscape. In addition, many existing models \textbf{cannot simulate vertical fire dynamics}, limiting their effectiveness during fast-evolving wildfire events \cite{xu2024wildfire, he2024study}. 


Building on this trajectory, emerging generative AI models—such as Variational Autoencoders (VAEs), Generative Adversarial Networks (GANs), and Transformers—offer substantial advantages over earlier deep learning architectures like CNNs and RNNs, particularly for complex environmental and urban modeling tasks \cite{bhatia1deep, zhang2024deep, lifelo2024artificial}. \rev{These models already power a wide range of applications in everyday software for personal use, content creation, and entertainment, including image synthesis, text generation, music composition, and interactive conversational systems, where they deliver capabilities far beyond those of conventional deep learning approaches \cite{li2025comprehensive}. Unlike traditional models that primarily specialize in classification or regression, generative AI systems learn intricate data distributions to produce entirely new, coherent, and contextually appropriate outputs \cite{sordo2025review}. This demonstrated ability to generate high-fidelity, multimodal content has sparked growing interest in adapting generative techniques to scientific and environmental domains, where similar benefits—such as robust uncertainty handling, scenario simulation, and the integration of heterogeneous data—hold significant transformative potential.}

\rev{Motivated by this potential, this paper investigates how the convergence of generative AI techniques can drive next-generation bushfire prediction systems. We begin by conducting a systematic review of recent studies that apply generative AI to fire prediction and monitoring, utilizing a new paradigm that leverages large language models (LLMs) for literature synthesis, classification, and knowledge extraction. Through this review, we examine how these technologies support \rev{high-fidelity spatial modeling and improved scalability and accuracy}, capabilities that are critical for enhancing situational awareness and decision-making during fire emergencies \cite{dilo2011data, ma2024generative}. We then outline five key future directions for advancing generative AI in wildfire management: developing unified multimodal frameworks that seamlessly integrate 2D and 3D data; designing conversational agentic AI systems for interactive, real-time wildfire intelligence; training interdisciplinary AI foundation models; enabling edge-based scenario generation on mobile and IoT devices; and advancing explainable AI interfaces to improve transparency and trust, accompanied by a discussion of critical challenges.
}

\section{Previous Reviews in Fire Spread Management}
\rev{Several previous studies have conducted comprehensive literature reviews on predicting bushfire spread using traditional methods, including simulation techniques, machine learning, and deep learning. While these approaches are not the primary focus of this paper—which centers on generative AI applications—they provide essential background on conventional fire spread prediction and its underlying scientific rationale.} 
This section begins by summarizing these reviews, highlighting how both simulation models and deep learning approaches have been applied to fire propagation prediction. The overview of previous review articles aims to provide a comprehensive understanding of the existing AI applications in wildfire spread prediction and to identify gaps where generative AI models can be leveraged to address challenges faced by traditional simulation and deep learning approaches.

\subsection{Wildfire Simulation Models and Traditional Machine Learning}
\label{sec:previous_lit_review}
Wildfire simulation models are traditional, well-established tools that simulate and predict bushfire spread by solving physical and empirical equations based on fuel, weather, and terrain conditions—originating from foundational models like Rothermel’s in the 1970s \cite{rothermel1972mathematical, finney2012need}, and they have long served as the backbone of operational fire management and planning systems. Based on their underlying rationale, these models can be classified into several categories, as presented in Table \ref{tab:simulation_summary}, and are summarized through multiple previous studies \cite{sullivan2009wildland, finney2012need}.

\begin{longtable}{p{2cm} p{3cm} p{3cm} p{3cm}}
    
    \caption{Fire simulation models based on different underlying principles} \\
    \toprule
    \textbf{Model Type} & \textbf{Core Principle} & \textbf{Examples} & \textbf{Strength}  \\
    \midrule
    \endfirsthead
    \endhead
    \bottomrule
    \endlastfoot

    \textbf{Physical Models}    
    & 
    Based on first principles of physics and chemistry (e.g., combustion thermodynamics, heat transfer, fluid dynamics)
    & FIRETEC, WFDS, FIRESTAR, IUSTI, Grishin
    & High fidelity, models full fire–fuel–atmosphere interaction, scientific rigor
    
    \\
    \textbf{Quasi-Physical Models}    
    & 
    Includes physical processes (e.g., energy conservation, heat transfer) but omits combustion chemistry, often uses simplified fire shape assumptions
    & UoS (Spain), LEMTA, FIRESTAR-lite
    & Balance between physical realism and computational feasibility
    
    \\
    
    \textbf{Empirical Models} 
    & Derived purely from statistical regression of observed fire behavior (no physical basis); field or lab-based
    & McArthur FDRS, CSIRO Grass Meter, CFBP (Canada)
    & Easy to use, computationally light, good for operational tools
    \\

    \textbf{Quasi-Empirical Models} 
    & Empirical models informed or supported by a physical framework (e.g., use physical insights to design empirical terms)
    & Rothermel model, BEHAVE, Noble-McArthur model
    & Widely used in practice, moderate complexity
    \\

    \textbf{Mathematical Analogue Models} 
    & Use abstract mathematical constructs (e.g., cellular automata, percolation theory, wavelet propagation) not rooted in real fire physics
    & Cellular Automata models, Huygens’ wavelet, Prometheus, SiroFire
    & Flexible, suitable for exploratory simulation, fast prototyping
    \\

\end{longtable}\label{tab:simulation_summary}

On the application level, a recent study conducts a comprehensive review of empirical and dynamic wildfire simulation models, focusing on their applications in predicting bushfire and wildfire spread across Australia \cite{singh2025comprehensive}. It critically evaluates a suite of simulation systems including PHOENIX Rapidfire, SPARK, AUSTRALIS, REDEYE, IGNITE, and SiroFire, each leveraging distinct modeling techniques to simulate fire behavior, together with a variety of traditional machine learning and deep learning methods for wildfire prediction. PHOENIX Rapidfire, a deterministic simulator, uses a fire characterization model based on Huygen's principle to predict spread, flame height, and intensity. While it excels in speed and ease of use, it is limited by reliance on predefined behavior models and underestimation of irregular fire shapes \cite{pugnet2013wildland, richards1995general}. SPARK provides flexible, modular simulation driven by user-defined spread models and integrates well with GIS platforms, though its high computational demands pose practical constraints. AUSTRALIS employs a discrete-event simulation based on empirically derived rate-of-spread models; despite its efficiency, it underperforms in severe fire conditions \cite{miller2015spark}. REDEYE and IGNITE integrate geospatial and real-time data for risk prediction and hotspot mapping, respectively, yet their effectiveness is limited by data compatibility and platform-specific requirements. SiroFire models dynamic weather components and supports strategic planning but lacks adaptability for diverse fuel types \cite{singh2025comprehensive}. Collectively, these simulators provide valuable predictive capabilities; however, each exhibits distinct limitations, ranging from computational inefficiencies and limited model adaptability to reduced accuracy under extreme conditions. This review highlights the need for hybrid modeling approaches that integrate traditional simulation techniques with machine learning and real-time data to enhance predictive accuracy, reduce simulation runtime, and strengthen operational resilience.

\subsection{Deep Learning in Wildfire Prediction}
\label{sec:previous_lit_review}

Several prior studies have conducted comprehensive reviews of machine learning models, including deep learning approaches, for wildfire prediction. In this work, we concentrate on the most recent literature review to provide an up-to-date overview of the deep learning models utilized and their respective application domains. As the application of machine learning and deep learning to wildfire prediction has already been extensively discussed and explored in numerous review papers over the past decades, this paper does not repeat those reviews. Instead, we focus specifically on reviewing, examining, and summarizing studies that apply emerging generative AI models, an area that has not been comprehensively covered in previous reviews.

In the following subsection, we leverage a LLM to generate a literature summary and bibliographic visualizations that highlight the evolving trends in deep learning applications over time. \cite{jain2020review} presents a comprehensive scoping review of machine learning (ML) applications in wildfire science and management, analyzing 300 studies published up to the end of 2019. ML usage is categorized across six core domains: fuel characterization, fire detection, climate interactions, risk prediction, fire behavior, and post-fire effects. Among modern approaches, deep learning (DL) algorithms have shown strong potential for modeling wildfire spread, particularly due to their capacity to handle high-dimensional and multivariate data. CNNs are widely used for spatial fire detection and smoke recognition from imagery, while Long Short-Term Memory (LSTM) networks incorporate temporal dynamics to model fire growth. Deep Neural Networks (DNNs) have also been applied to map burned areas and forecast fire spread, often outperforming traditional models when sufficient annotated data are available. Despite their promise, the review emphasizes the importance of domain expertise and high-quality data for meaningful DL integration in operational forecasting.

A recent study systematically reviews deep learning (DL) applications in forest fire prediction, analyzing 55 key publications from 2017 to 2024 \cite{mambile2024application}. The review emphasizes the transformative role of deep learning in capturing complex spatiotemporal patterns associated with fire ignition and spread, offering significant advantages over traditional machine learning approaches. CNNs are effectively used for analyzing satellite imagery to detect fire-prone areas and assess post-fire damage. LSTM and Gated Recurrent Unit (GRU) networks model time-series data to forecast fire progression based on historical environmental conditions. Emerging generative AI models such as GANs have been used for synthetic data generation, while Multilayer Perceptrons (MLPs) and U-Net architectures support fire risk estimation and boundary segmentation. Despite promising results, DL models face challenges such as data heterogeneity, limited inclusion of human activity factors, generalization across geographies, and the scarcity of high-quality labeled datasets. These limitations highlight the need for integrating diverse data sources and establishing standardized evaluation protocols to ensure model reliability in real-world wildfire prediction scenarios.

\subsection{Limitations of the Existing Deep Learning Applications in Wildfire Prediction}
\label{sec:limit_dl_app}
Based on a comprehensive review of existing literature at the intersection of deep learning and wildfire prediction, as well as an analysis of the inherent challenges and limitations of various deep learning models stemming from their underlying mechanisms and data-driven assumptions, we identify several critical knowledge gaps and challenges that constrain the effectiveness and broader applicability of current deep learning approaches in wildfire prediction.

\begin{description}
    \item[L1. Quantification of Limited Uncertainty:] Traditional models such as CNNs and RNNs typically produce deterministic outputs and struggle to quantify prediction uncertainty, an essential capability for wildfire management and emergency response applications involving stochastic environmental processes, such as abrupt changes in wind speed and directions \cite{gal2016dropout,lakshminarayanan2017simple}.
 
    
    \item[L2. Weak Long-Term Dependency Modeling:] Recurrent Neural Networks (RNNs) and Deep Neural Networks (DNNs) often struggle with vanishing gradients and limited temporal memory, which undermines their ability to capture long-range dependencies. This limitation makes them less effective in modeling the temporal progression of wildfires and in representing the long-term variability of underlying environmental processes such as fuel accumulation, climate patterns, and vegetation dynamics \cite{hochreiter2001gradient, chen2024explainable}.

    
    \item[L3. Inadequate Multimodal Data Integration and Prediction:] Traditional deep learning architectures are not inherently designed to fuse multimodal data sources, such as 2D GIS data (e.g., satellite imagery, fuel load maps, and meteorological data) and 3D point clouds (e.g., digital terrain models) \cite{li2022deep}. As a result, generating multimodal fire spread prediction outputs across different dimensions—from 1D time series to 3D spatial representations—within a unified deep learning framework remains a significant challenge.

     
    \item[L4. Limited Data Augmentation Capabilities:] Traditional deep learning models, such as CNNs and RNNs, typically achieve strong performance only when trained on abundant, high-quality labeled data. They depend heavily on large volumes of annotated samples, which are often scarce in wildfire prediction tasks due to the rare, unpredictable, and spatially heterogeneous nature of fire events \cite{ghali2023deep, xu2024wildfire}. However, many of these models lack the capability to perform data augmentation or generate synthetic training samples, limiting their predictive accuracy and generalizability in data-sparse or unseen regions. This limitation poses a significant challenge for developing robust wildfire.

    
    \item[L5. Missing Data and Poor Data Quality Challenges:] Environmental datasets frequently contain missing or incomplete information caused by sensor malfunctions, occlusion from cloud cover in satellite imagery, or data transmission failures \cite{shen2015missing}. Such data gaps hinder accurate modeling and prediction. Traditional deep learning models often assume complete input data or rely on simplistic imputation methods that fail to capture the underlying spatiotemporal dependencies critical for wildfire dynamics. These limitations reduce model robustness and prediction accuracy in real-world scenarios where data is inherently noisy or sparse \cite{li2020misgan}.


    
    
    \item[L6. Lack of Explainability:] Deep learning models such as CNNs and RNNs often operate as "black boxes," providing limited insight into how predictions are made and offering little transparency or trustworthiness \cite{rudin2019stop, doshi2017towards}. 
    
\end{description}

While previous studies have provided structured and comprehensive reviews, most recent research remains focused on traditional machine learning and deep learning methods, or at most covers only a specific type of generative AI model for wildfire prediction. In contrast, comprehensive reviews and in-depth discussions that encompass the broader spectrum of emerging generative AI models, along with explanations of their underlying algorithmic principles in this domain, remain scarce and largely underexplored.

In the following section, we examine recent emerging studies that apply generative AI models to wildfire spread prediction, discussing the underlying rationale for using generative approaches and highlighting their advantages over traditional deep learning methods in addressing the aforementioned limitations. Our discussion then expands to explore how popular generative AI models can outperform conventional deep learning approaches in certain predictive analytics tasks. Ultimately, we aim to offer new perspectives on the potential of generative AI in next-generation emergency response systems, enabling faster, more reliable, and higher-resolution predictions of wildfire spread.

\section{Emerging Generative AI Models and Their Advantages}

Since around 2018 and especially in the early 2020s, generative AI models have experienced rapid advancements and widespread adoption across diverse fields, marking a significant shift in the landscape of artificial intelligence \cite{brown2020language}. Broadly defined, generative AI refers to a class of machine learning models that are capable of learning the underlying distribution of data and generating new, realistic content—such as text, images, audio, or even simulations—that resembles the training data \cite{goodfellow2014generative, kingma2013auto, ho2020denoising}. These models and their architectures have been increasingly experimented with and adopted to solve complex problems across various scientific disciplines. 

\subsection{Generative AI Applications in Environmental and Urban Sciences}
The recent emerging generative AI models go beyond traditional predictive analytics by creating novel outputs, making them highly valuable in data-scarce environments or in domains requiring scenario generation, synthesis, or simulation. Generative AI encompasses several prominent categories, including GANs, VAEs, Autoregressive Models (AR), Diffusion Models, and Flow-based Models \cite{vaswani2017attention, brown2020language,kingma2013auto}. Within these categories, more specialized types have emerged: GANs consist of a generator and discriminator in adversarial training to synthesize realistic data \cite{goodfellow2014generative}; VAEs combine probabilistic encoding and decoding for efficient latent space learning \cite{kingma2013auto}; autoregressive models like GPT-4 predict future tokens in sequences for high-quality text generation \cite{brown2020language}; diffusion models such as Stable Diffusion and denoising diffusion probabilistic models (DDPMs) generate content by gradually denoising random noise into coherent samples \cite{ho2020denoising, rombach2022high}; and flow-based models leverage invertible transformations for exact likelihood estimation and sample generation \cite{dinh2017density}. 


Each generative AI model and its architecture offers distinct mathematical properties and advantages depending on the application context. The rise of these models has revolutionized data science and AI by enabling machines to not only interpret and predict but also to imagine and create, thereby unlocking new opportunities across fields including hazard prediction, urban planning, climate science, decision support, design automation, and digital twin construction \cite{xu2024leveraging,ma2024generative}. In the environmental hazard domain, \cite{ma2024generative} presents a comprehensive review of how generative AI addresses longstanding challenges in data availability, quality, and resolution. These models learn high-dimensional probability distributions from limited samples, making them ideal for data-scarce applications such as geohazards, hydrometeorology, and climate-related analysis. By generating physically consistent synthetic data—ranging from downscaled meteorological fields to simulated landslide and seismic imagery—GenAI enhances forecasting accuracy, susceptibility mapping, and early warning systems, driving more reliable and scalable hazard modeling frameworks. In the urban management sector, \cite{xu2024leveraging} conducts a scoping review on integrating GenAI into urban digital twins, highlighting its transformative role in automating the generation of high-quality urban data, hypothetical planning scenarios, and 3D city models. These capabilities help overcome major challenges related to data sparsity, simulation scalability, and design complexity. Through the synthesis of multi-modal urban data and simulation of complex dynamics, GenAI enhances real-time decision support, predictive analytics, and participatory planning across sectors such as transportation, energy, water, and infrastructure. The convergence of GenAI with digital twin technologies represents a paradigm shift toward intelligent, adaptive, and inclusive urban solutions. Furthermore, growing research extends GenAI applications into other urban subsystems, including logistics optimization, intelligent transportation systems, and domain-specific knowledge generation \cite{tupayachi2024towards, xu2024genai, taiwo2025making}.

\subsection{\rev{Theoretical Foundations of Fire Spread Modeling for Generative AI Integration}} \label{sec:new_opportunities}

\rev{Generative AI provides powerful computational tools to simulate, model, and predict natural phenomena, but it still relies on existing physical laws and fundamental principles. To explore how emerging generative AI architectures can enhance or redefine current wildfire spread simulations and their core processes, it is crucial to first revisit the fundamental principles underlying fire spread modeling.
At their core, most wildfire spread simulators are built upon two foundational computational paradigms: Huygens’ principle-based wavefront expansion methods and grid-based local interaction models. Each embodies a distinct conceptual strategy to replicate the spatiotemporal dynamics of fire growth, offering complementary strengths and insights for different use cases \cite{sun2023facing}.
} 

\rev{Huygens’ principle-based methods simulate wildfire spread by borrowing ideas from wave physics, focusing on how the physical process propagates over time. They treat every point along the fire edge as a new ignition point that pushes the fire outward, often forming an ellipse shaped by wind, slope, and fuel \cite{anderson1982modelling}. This helps capture how fires stretch and move in different directions. Many widely used systems, like Prometheus, FARSITE, and FlamMap, rely on this approach to keep updating the fire boundary over time \cite{tymstra2010development, mitsopoulos2014mapping}. 
On the other hand, grid-based models focus on the characterization and connectivity of individual spatial or volumetric units created by dividing the landscape into a grid of small squares (2D cells) or cubes (3D voxels). This grid provides a spatial and temporal framework over which physical processes, such as wildfire, can propagate \cite{barton2020voxel}. Each unit records its combustion state—unburned, burning, or burned—and captures local fuel and environmental characteristics. Fire spreads by following local transition rules influenced by fuel type, wind, slope, and stochastic effects \cite{sun2023facing}. Tools like Cell2Fire and hybrid models that combine grid methods with physics or machine learning build on this idea \cite{rui2018forest,xu2022modeling}. While often exploratory, grid-based models are valuable for studying detailed fire patterns, testing new hypotheses, and extending simulations into 3D to better capture fire behavior.
}
 
\rev{From a computational standpoint, generating large-scale grid-based models often requires substantial resources, including high RAM and GPU memory. Initializing large, connected grids in parallel environments can also incur significant time costs due to communication and load-balancing overhead. Moreover, both Huygens’ principle-based and grid-based methods typically depend on integrating multiple layers of GIS and meteorological data—such as topography, fuel attributes, and weather conditions—to achieve accurate physical or empirical simulations. Consequently, simulating wildfire spread over large areas or extended timeframes becomes highly resource-intensive, significantly increasing inference time and limiting the practicality of these approaches for rapid or wide-area fire behavior forecasting.
}

\rev{
To provide a clearer theoretical foundation regarding how these generative approaches can be integrated into the Huygens’ principle-based methods and the grid-based model, we first outline the core mathematical formulations that govern their learning objectives and elucidate their ability to capture both local and non-local fire spread dynamics. 
Conceptually, VAEs, diffusion models, and Transformers can be seamlessly integrated with these core fire spread simulation principles by operating over localized kernels of neighboring cells surrounding active burning fronts. This architecture enables the models to learn spatial and temporal transitions of adjacent unburned cells, effectively guided by the propagation dynamics derived from physics-based simulation outputs or real-world data. Simultaneously, by embedding fuel characteristics and environmental covariates within their latent representations or as embeddings, these models capture the complex, multifactorial behavior underlying wildfire propagation, thus offering a more computationally efficient and rapid means to model and predict fire spread beyond traditional physical rules \cite{yang2024survey, joshi2025introduction}.}

\rev{
Conceptually, VAEs optimize a stochastic latent-variable model by maximizing the evidence lower bound (ELBO), which balances reconstruction accuracy against divergence from a prior distribution and thus enables learning complex distributions over fine-scale spatial fire spread patterns. In contrast, diffusion models learn to reverse a gradual noising process through a denoising objective, allowing them to generate plausible next-timestep fire perimeters defined by Huygens’ principle and conditioned on environmental factors \cite{zhihan2024exploring}. Meanwhile, transformers use self-attention mechanisms to dynamically capture long-range spatial and temporal dependencies, making them well-suited to learn how fires propagate across heterogeneous landscapes without explicitly encoding all physical interactions. In the following section, we examine the mathematical formulations of each generative AI architecture and explore, at a theoretical level, how they can be integrated with the core principles of fire spread prediction.
}

\subsubsection{\rev{VAE}}
\rev{
Diving into the details, VAEs optimize a stochastic latent-variable model by maximizing the ELBO, which is represented in the following equation that balances reconstruction fidelity with regularization toward a prior, enabling VAEs to have the potential ability learn distributions over plausible local fire spread patterns \cite{kingma2014auto}. 
\begin{align}
\mathcal{L}_{\text{VAE}} 
= \mathbb{E}_{q_\phi(\mathbf{z}|\mathbf{x})}\!\big[\log p_\theta(\mathbf{x}|\mathbf{z})\big] 
- D_{\mathrm{KL}}\!\big(q_\phi(\mathbf{z}|\mathbf{x}) \,\|\, p(\mathbf{z})\big),
\end{align}
\label{eq:1}
In the equation \ref{eq:1}, $x$ represents the observed data—such as spatial snapshots of wildfire perimeters or voxel grids showing burned states—while $z$ denotes latent variables that capture hidden influences like wind, fuel, and terrain effects on fire spread. The encoder $q_\phi(z|x)$, parameterized by $\phi$, approximates the posterior distribution over these latent factors given the observed fire data, and the decoder $p_\theta(x|z)$, parameterized by $\theta$, reconstructs or generates fire spread patterns from them. The model optimizes the evidence lower bound (ELBO), balancing the reconstruction term $\mathbb{E}_{q_\phi(z|x)} [\log p_\theta(x|z)]$, which ensures the decoded spread closely matches the observed data, and the KL divergence term $D_{KL}(q_\phi(z|x) \| p(z))$, which regularizes the latent space toward a simple prior $p(z)$. In wildfire applications, this enables VAEs to learn meaningful compressed representations of local fire dynamics and to generate realistic alternative spread scenarios that remain consistent with underlying physical and environmental processes.
}

\subsubsection{\rev{Diffusion Models}}
\rev{
Diffusion models learn to reverse a gradual noising process, often through a simplified denoising objective (as shown in equation \ref{eq:2}), allowing them to have the ability to generate realistic next-step fire perimeters conditioned on environmental features \cite{ho2020denoising}. 
\begin{align}
\mathcal{L}_{\text{DM}} 
= \mathbb{E}_{\mathbf{x},\epsilon,t} 
\!\!\left[ 
\big\| \epsilon - \epsilon_\theta(\mathbf{x}_t, t) \big\|^2 
\right],
\end{align}
\label{eq:2}
In this equation, $x$ denotes the original data sample, which in wildfire modeling could represent a spatial map or voxel grid of a current fire perimeter or fuel state. The variable $t$ indexes the timestep within the diffusion process, tracking the progression of noise addition or removal. Correspondingly, $x_t$ is the noisy version of the data at step $t$, simulating partially corrupted fire perimeter states. The term $\epsilon$ represents the actual noise sampled and applied during the forward diffusion process, while $\epsilon_\theta(x_t, t)$ is the model's prediction of this noise, parameterized by $\theta$. By minimizing the difference between the true noise and the predicted noise across all samples, noise levels, and timesteps—captured by the expectation $\mathbb{E}_{x,\epsilon,t}$—the model learns to effectively reverse the noising process. In wildfire applications, this can potentially enable the generation of realistic next-timestamp fire perimeters conditioned on environmental features, by iteratively refining uncertain or noisy predictions into plausible spread scenarios that align with learned data distributions and underlying physical patterns.
}

\subsubsection{\rev{Transformers}}
\rev{
Finally, transformers use self-attention to model long-range spatial or temporal dependencies (as depicted in equation \ref{eq:3}), which enables dynamic context aggregation across heterogeneous landscapes or multi-scale temporal sequences \cite{vaswani2017attention}. 
\begin{align}
\text{Attention}(Q,K,V) 
= \text{softmax}\!\!\left(\frac{QK^\top}{\sqrt{d_k}}\right) V,
\end{align}
\label{eq:3}
In this equation, $Q$, $K$, and $V$ represent the query, key, and value matrices, respectively. These are learned projections of the input data that capture different spatial or temporal features. In wildfire spread modeling, for example, $Q$ might encode the characteristics of a focal location or time step—such as current fuel moisture, wind conditions, or ignition state—while $K$ and $V$ capture information from surrounding spatial cells or earlier temporal states. The operation $\mathrm{softmax}!\left(\frac{QK^\top}{\sqrt{d_k}}\right)$ computes attention weights by measuring the similarity between the query and the keys, normalized by the dimension $d_k$ to ensure stable gradients. These weights then aggregate the values $V$, producing a context-aware representation that dynamically integrates long-range dependencies. In the context of fire spread, this mechanism could be incorporated into a grid-based framework to learn how a cell’s or voxel’s future state is influenced by conditions across heterogeneous landscapes and multi-step temporal interactions, effectively complementing classical Huygens’ principle and traditional grid-based simulators by automatically capturing complex, data-driven spatial and temporal relationships.
}
 \begin{table}[htbp]
\centering
\caption{Comparison of Traditional Machine Learning, Deep Learning, and Generative AI models.}
\begin{tabular}{|p{2cm}|p{3cm}|p{3cm}|p{4cm}|}
\hline
\textbf{Feature} & \textbf{Traditional ML} (e.g., SVM, RF) & \textbf{Deep Learning} (e.g., CNN, LSTM) & \textbf{Generative Deep Learning} (e.g., VAE, GAN, Transformer, Diffusion) \\
\hline
Data Generation Capability & Cannot generate data & Predictive only &  Can generate new, realistic, diverse data \\
\hline
Representation Learning & Manual feature engineering & Learns hierarchical features & Learns latent and semantic representations \\
\hline
Handling Missing Data / Imputation & Basic imputation (e.g., mean) & Regression-based or interpolation only & Learns to impute based on data distribution \\
\hline
Few-shot / Zero-shot Learning & Requires full training & Requires full training & Supported by large-scale transformers \\
\hline
Multimodal Learning & Needs manual integration 
& Separate networks for each modality 
& Unified models handle text, image, video, etc. \\
\hline
Uncertainty Quantification & Via Bayesian methods or ensembles & Deterministic output & Built-in probabilistic frameworks (e.g., VAEs) \\
\hline
Synthetic Data Augmentation & Not supported & Requires manual engineering & Easily supports realistic data generation \\
\hline
Scenario Simulation & Not applicable & Not applicable & Simulates realistic and hypothetical conditions \\
\hline
Latent Space Manipulation & Not available & Not interpretable & Supports interpolation and control \\
\hline
Creativity \& Generative Power & None & None & High—generates novel outputs and scenario \\
\hline
Data Efficiency & Needs many labeled samples & High data demand & Some support few-shot learning via pretraining \\
\hline
Interpretability & Often interpretable & Difficult to interpret hidden layers & Latent space can be visualized and interpretable \\
\hline
Training Complexity & Simple to train & Needs tuning and GPU support & Complex training and high computational cost \\
\hline
Use in Scientific Simulation & Limited (basic regression) & Used in some modeling & Strong in data-driven and uncertain modeling \\
\hline
\end{tabular}
\label{tab:ml_dl_genai_comparison}
\end{table}

\subsection{Advantage over Traditional Deep Learning Models}
Building on the limitations identified in the existing wildfire prediction literature, we conducted a comparative analysis of traditional deep learning models and generative AI approaches. \rev{As summarized in Section~\ref{sec:limit_dl_app}, our findings highlight the key advantages of generative AI models in overcoming the limitations of conventional methods. These advantages are outlined below, with a detailed comparison presented in Table~\ref{tab:ml_dl_genai_comparison}, which maps each limitation (L1–L6) discussed in the previous section to the corresponding strengths of generative AI.}
 
\begin{description}
    \item[Richer Uncertainty Modeling:] In contrast to many traditional deep learning models, generative AI models such as VAEs and diffusion models produce probabilistic outputs that inherently capture uncertainty in predictions—an essential capability for high-risk applications like wildfire forecasting \cite{kingma2013auto, ho2020denoising}. This feature can be leveraged to address the limitation L1.

    \item[Better Long-Term Dependencies:] Transformer-based architectures outperform recurrent neural networks (RNNs) in modeling long-range dependencies in sequential data, making them well-suited for tracking wildfire dynamics over extended time periods \cite{vaswani2017attention, brown2020language} The capability could be harnessed to tackle the limitation L2.

    \item[Multimodal Data Fusion:] Generative AI models, particularly those based on transformers and diffusion techniques, excel at integrating heterogeneous data sources (e.g., satellite imagery, meteorological variables, and point clouds), enabling more robust 2D and 3D wildfire forecasting through a unified framework \cite{rombach2022high, radford2021learning} to address the limitation L3.

    \item[Data Augmentation \& Synthesis:] Generative AI models such as GANs and diffusion models can produce synthetic yet realistic wildfire progression data, providing valuable training samples in data-scarce scenarios, facilitating missing data imputation to enhance data quality, and supporting the simulation of extreme or hypothetical conditions \cite{goodfellow2014generative, dhariwal2021diffusion}. This capability can be leveraged to address Limitations L4 and L5, as well as to enable scenario generation for simulating hypothetical wildfire events.
    
    \item[AI Explainability through Latent Space:] Many generative AI models, such as VAEs and GANs, rely on latent spaces and latent vectors to operate, where complex data relationships are encoded in lower-dimensional representations \cite{kingma2013auto, goodfellow2014generative}. These latent variables can be visualized and analyzed to enhance the explainability, interpretability, and trustworthiness of the models in tasks such as classification, prediction, clustering, and data generation \cite{xu2024explainable}, thereby addressing Limitation L6.

\end{description}

The unique advantages of generative AI models are systematically inferred from their architectural principles and algorithmic rationale, particularly in the context of data-driven wildfire prediction. Applications of these models often rely on standard environmental and urban GIS datasets—such as shapefiles, raster imagery, and 3D LiDAR point clouds—which are also widely used in generative AI research across other domains. In Section~\ref{sec:review_fire_genAI}, we present a comprehensive and critical review of studies applying generative AI to wildfire prediction, highlighting their strengths for bushfire management. Our subsequent review of generative AI models aims to explore future opportunities for leveraging their strengths to revolutionize bushfire prediction, as well as to discuss the associated challenges. The insights gained from this review are presented in Section~\ref{sec:new_opportunities}.

\section{Review Strategy}

\begin{figure}[H]
    \centering
    \includegraphics[width=1\textwidth]{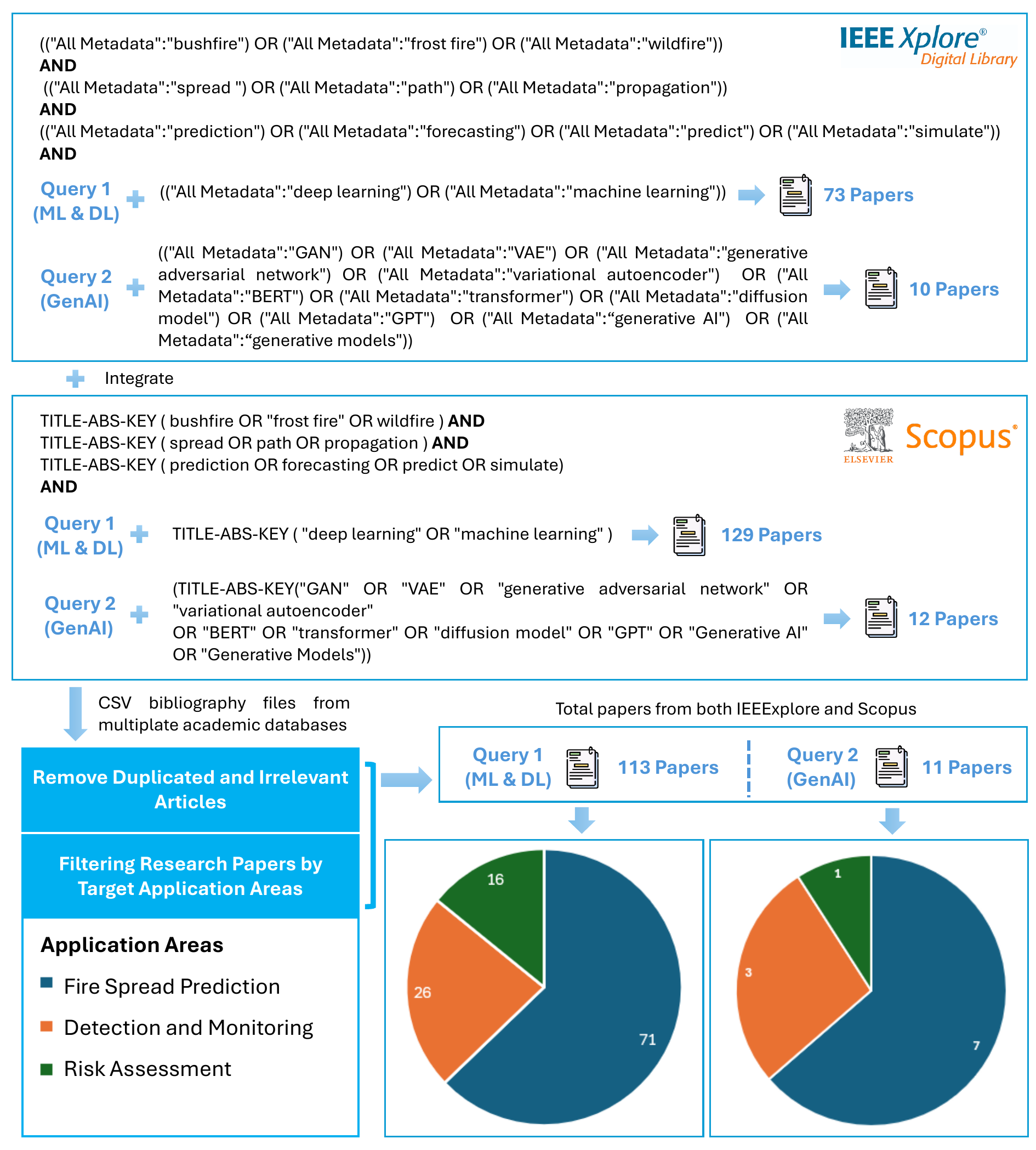}
    \caption{Search queries used to identify and acquire literature from IEEExplore and Scopus databases.}
    \label{fig:literatyre_search}
\end{figure}

\rev{To ensure methodological rigor and transparency, this study adopts a systematic review strategy combined with narrative synthesis. By systematically formulating search queries and applying them across well-established academic databases—including IEEE Xplore and Scopus—we identified and screened relevant literature in a structured, reproducible manner (as depicted in Figure~\ref{fig:literatyre_search}). This approach enabled us to comprehensively map existing research on the application of artificial intelligence to wildfire spread prediction, while also allowing us to iteratively narrow our scope to focus on the emerging role of generative AI models. Following initial retrieval and screening, we employed a narrative synthesis to qualitatively analyze and compare selected studies, examining their methodological frameworks, data sources, and reported predictive performances. This process provided nuanced insights into how diverse deep learning and generative architectures have been leveraged to advance fire propagation modeling.}

\rev{
In addition to the identified articles, we employed an LLM-powered tool from our previous work \cite{xu2024automating} to characterize the literature by extracting taxonomies from each paper’s abstract using a combined process of Scientific Discourse Tagging (SDT) and Named Entity Recognition (NER). These taxonomies were then classified against an existing body of knowledge. Specifically, we used Sentence Transformers to align each article’s extracted taxonomy with domain knowledge by applying a cosine similarity threshold, comparing entities to definitions from a domain ontology built through comprehensive literature reviews and authoritative textbooks. A similarity threshold of 0.7 was chosen to ensure robust alignment, classifying papers under specific methodologies or application areas only when their extracted taxonomy showed high semantic relevance to existing definitions. For example, this process enabled classification of literature into application areas such as fire spread prediction, detection and monitoring, and risk assessment, as illustrated in Figure~\ref{fig:literatyre_search}.
}
 
\section{Generative AI Applications in Wildfire Management}
\label{sec:review_fire_genAI}

Through the following subsections, we present a structured, in-depth review of 11 existing studies that have explored the use of generative AI models for wildfire management. \rev{This review is organized according to the application areas shown in Figure~\ref{fig:literatyre_search}, which were identified through keyword-based literature characterization. Our primary aim is to examine how generative AI has been applied to support fire prediction, while also highlighting related areas such as fire monitoring and risk mapping. We focus specifically on evaluating the prediction accuracy, inference time, and computational efficiency of these generative AI models compared to traditional methods used as benchmarks.}

\begin{figure}[H]
    \centering
    \includegraphics[width=1\textwidth]{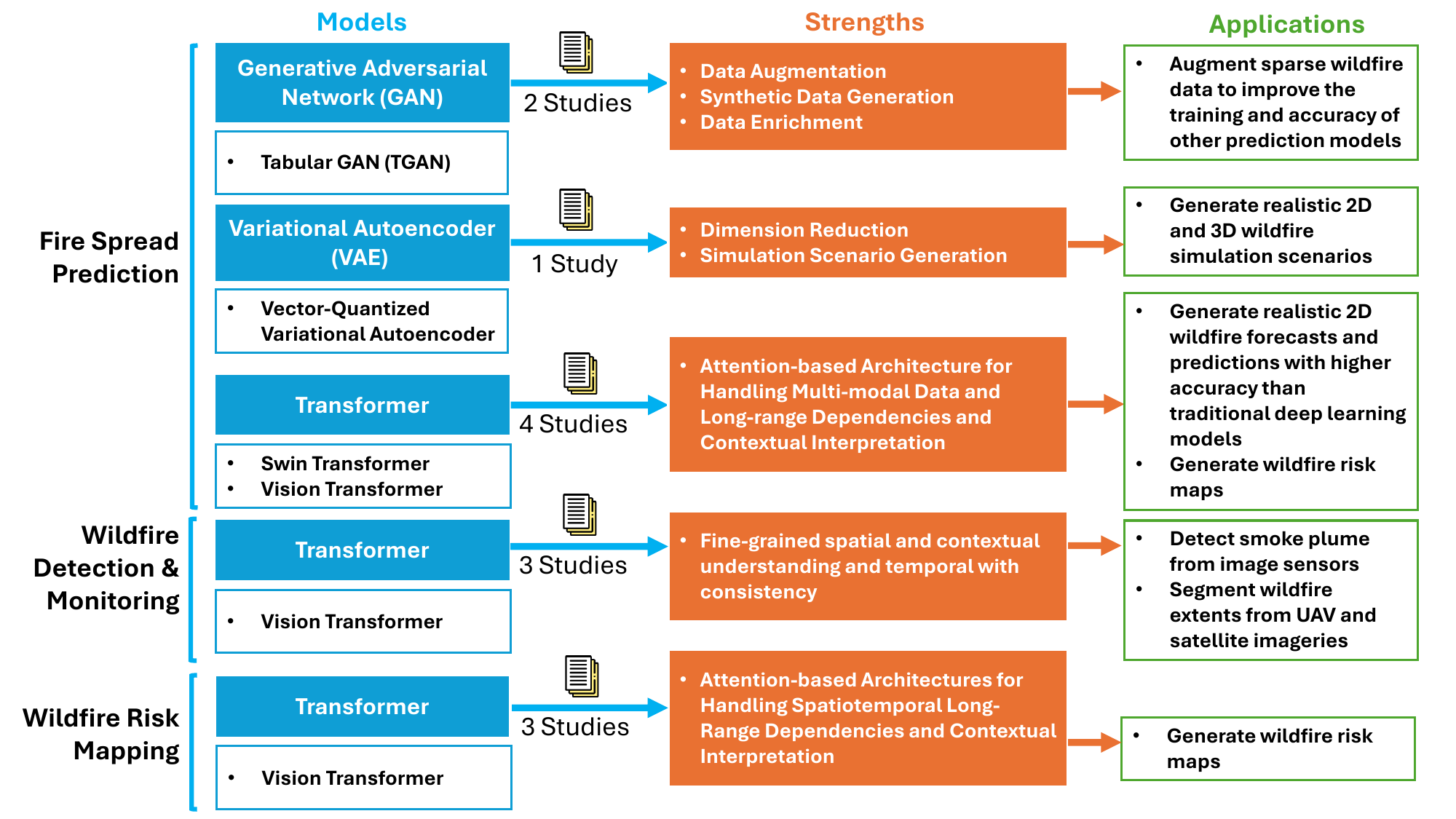}
    \caption{Summary of recent studies applying generative AI to diverse areas in bushfire modeling and prediction.}
    \label{fig:genAI_literature_summary}
\end{figure}

\subsection{Fire Spread Prediction}
We identified seven existing studies that have explored or prototyped the use of various generative AI models to support fire spread prediction and fire behavior modeling. These studies leverage \rev{three} different architectures, \rev{namely} GANs, VAEs, and generative transformers, to simulate wildfire dynamics, generate synthetic fire scenarios, or enhance predictive performance. Collectively, they demonstrate the emerging potential of generative AI in capturing the complex spatiotemporal patterns of wildfire propagation and improving the accuracy, scalability, and adaptability of fire behavior forecasting systems.
 
\subsubsection{Generative Adversarial Network (GAN)}
GANs and their variants have proven valuable in environmental and urban research by generating synthetic data to augment limited and imbalanced datasets \cite{xu2024leveraging}. This capability is especially critical in wildfire research, where collecting real-world fire data is often constrained by safety risks, high costs, and logistical challenges. By simulating underrepresented fire scenarios—such as passive and active crown fires—GANs help improve the diversity and robustness of training data, leading to more accurate classification of fire behavior and better support for operational decision-making. We identified two studies that effectively employ GANs for data augmentation to enhance wildfire spread prediction through integration with machine learning and simulation models.

\cite{khanmohammadi2023artificial} presents a machine learning framework for predicting wildfire spread sustainability and crown fire occurrence in semiarid shrublands of southern Australia. The study uses a TGAN to generate synthetic fire records, addressing the challenge of small training datasets commonly encountered in this domain. The inclusion of synthetic data led to substantial performance improvements, increasing classification accuracy by up to 20\% for spread sustainability and 4\% for crown fire prediction. This enhanced dataset was used to train and evaluate several classifiers—including Support Vector Machines (SVM), Multilayered Neural Networks (MLP), and Multinomial Naive Bayes (MNB)—with the SVM model paired with TGAN-generated data achieving the highest accuracy. The results demonstrate the potential of combining generative AI with traditional supervised learning models to improve generalization and reliability in data-scarce wildfire prediction settings. \rev{Trained on a dataset comprising 61 experimental fires in Australian semiarid shrublands—augmented by 27 TGAN-generated samples for spread sustainability and 13 for crown fire cases—the TGAN-enhanced SVM achieved 90\% accuracy for predicting sustained fire propagation and 80\% accuracy for active crown fire occurrence on an independent evaluation set of 29 fires. Compared to logistic regression models developed on the same data, this represented improvements of 15 percentage points for spread sustainability and 4 points for crown fire prediction, along with markedly higher sensitivities (1.0 vs. 0.77 and 0.55, respectively). Although the study did not report training or inference times, it clearly demonstrated that integrating TGAN-based data augmentation substantially boosts predictive performance and robustness relative to traditional statistical approaches, especially under data-limited conditions.}

\cite{khanmohammadi2024using} extends this approach to Canadian conifer forests, focusing on predicting wildfire propagation types, including surface, passive crown, and active crown fires. In this study, TGAN is again employed to address class imbalance by generating synthetic data for underrepresented fire types, significantly boosting prediction accuracy and F1-scores—especially for difficult-to-classify categories like passive crown fires. The study evaluates four additional machine learning models: two ensemble methods (Random Forest and XGBoost) and two AutoML approaches (TabPFN and AutoGluon), with TabPFN consistently achieving the best results when trained on GAN-augmented data. These findings reinforce the value of GAN-based data generation as a means of overcoming real-world data limitations and enhancing the predictive capabilities of wildfire behavior models. \rev{Trained on a dataset of 113 experimental fire cases collected over 60 years in Canadian conifer forests, including 52 surface fires, 23 passive crown, and 38 active crown fires, the study demonstrated that using TGAN to generate synthetic records dramatically improved prediction performance. With balanced datasets created by adding up to 26 synthetic passive crown and 13 active crown fire samples, the best generative AI-related model, TabPFN, achieved 91\% accuracy and a 93\% F1-score on independent evaluation data in the binary classification task, outperforming traditional logistic regression baselines. In the more challenging multi-class setting (surface vs. passive crown vs. active crown), the addition of GAN data elevated TabPFN’s accuracy from 61\% to 74\% and its F1-score from 50\% to 65\%, particularly improving predictions of passive crown fires. Although exact training or inference times were not provided, the study clearly highlighted how TGAN-based data augmentation substantially enhances the spatial-temporal prediction fidelity of wildfire models under data-limited scenarios.}

\rev{In Summary, previous studies have shown that integrating GAN-based data augmentation can substantially outperform traditional simulation and machine learning approaches in fire spread prediction and behavior modeling \cite{khanmohammadi2023artificial, khanmohammadi2024using}. By leveraging TGANs to generate synthetic wildfire records, these works effectively addressed class imbalance and limited data, achieving up to 90–91\% accuracy and F1-scores exceeding 93\%, notably outperforming logistic regression baselines. In more complex multi-class settings involving surface, passive crown, and active crown fires, GAN-augmented models improved accuracy by over 13 percentage points and F1-scores by 15 points, significantly enhancing the detection of underrepresented fire types. These findings highlight the strength of GAN-driven data generation in boosting the fidelity, robustness, and generalizability of wildfire forecasting models beyond what conventional statistical or standard deep learning methods can achieve under data-scarce conditions.}

\subsubsection{Variational autoencoder (VAE)}
The use of VAEs and their variants is emerging in the wildfire prediction sector to generate realistic spatiotemporal simulations of fire spread, serving as synthetic training data for downstream forecasting models. These generative models are particularly effective for tasks that require the synthesis of physically consistent fire evolution sequences from high-dimensional, multi-modal inputs, such as topography, vegetation density, and weather conditions, where real data may be scarce, incomplete, or expensive to obtain through traditional physics-based simulations.

\rev{\cite{cheng2023generative} introduces a generative AI framework to overcome the computational demands of traditional physics-based wildfire simulations. Using a Vector-Quantized Variational Autoencoder (VQ-VAE) trained on Cellular Automata (CA) data, the model generates high-fidelity 3D sequences of wildfire spread that capture essential geophysical influences like vegetation and slope. This approach accelerates data generation by over four orders of magnitude compared to CA and MTT simulators. The synthetic wildfire scenarios are then used to train a POD-LSTM surrogate model, which, with the inclusion of VQ-VAE data, achieves notably higher prediction accuracy and structural similarity on both simulated and real events. This demonstrates the VQ-VAE’s effectiveness in reducing simulation costs, enriching training datasets, and improving the realism and generalizability of wildfire forecasts. Specifically, the VQ-VAE was trained on 40 CA-generated fire spread sequences from the Chimney Fire region in California, each spanning 8 days (16 temporal snapshots) at 128×128 spatial resolution, and was then used to generate 500 additional synthetic fire scenarios. This resulted in surrogate models trained on VQ-VAE–augmented data achieving significantly lower relative RMSE and higher structural similarity index (SSIM) than models trained solely on the limited CA data, for both unseen synthetic fires and the real Chimney Fire event observed via MODIS and VIIRS satellites. In terms of computational efficiency, the VQ-VAE generated 8-day wildfire spread sequences in just 0.3 seconds, representing a speed-up of 4–5 orders of magnitude compared to traditional CA and MTT simulators, effectively overcoming the major runtime bottlenecks of physics-based wildfire modeling while preserving crucial spatial-temporal dynamics.}


\subsubsection{Transformer}
Transformers have emerged as powerful deep learning architectures for wildfire prediction tasks, particularly in modeling fire spread, classifying risk levels, and generating predictive spatial outputs from complex spatiotemporal data. Their capacity to capture long-range dependencies and contextual interactions across diverse input modalities, such as satellite imagery, weather data, topography, and vegetation maps—makes them especially well, suited for wildfire forecasting tasks that require spatial precision and temporal consistency to support timely decision-making and emergency response.

\cite{li2024wildfire} presents a transformer-based hybrid model—Attention Swin U-Net with Focal Modulation (ASUFM)—designed for next-day wildfire spread forecasting across North America. ASUFM integrates spatial attention and focal modulation layers into a Swin Transformer U-Net backbone, producing predictive fire masks from multivariate remote sensing data. Though not a conventional generative model, ASUFM simulates realistic fire spread scenarios and addresses challenges such as spatial resolution, class imbalance, and temporal consistency. Trained on an expanded NDWS dataset (2012–2023), the model achieved state-of-the-art results in Dice score and PR-AUC, outperforming U-Net and other transformer variants. Additional techniques, including weighted loss functions, skip connections, and cosine learning rate scheduling, further enhanced the model's generalization and accuracy. \rev{On the extended NDWS dataset covering over 31,000 samples from 2012–2023 across all of North America at 1 km spatial resolution, ASUFM achieved a Dice score of 0.4066, precision of 0.4345, recall of 0.4096, and a PR-AUC of 0.3974, substantially outperforming traditional encoder-decoder baselines such as U-Net (Dice 0.3493, PR-AUC 0.2945) and even pure Swin U-Net variants. This highlights the transformer’s superior ability to model long-range dependencies and capture complex spatial fire propagation patterns. Although the study did not report explicit training or inference times, it noted using large-scale GPU infrastructure (NVIDIA A100-80G or RTX A6000) with advanced techniques like focal modulation and spatial attention to balance precision-recall under heavy class imbalance. The expanded spatial and temporal coverage of the dataset ensures that ASUFM’s learned representations generalize across diverse geographic terrains and multi-year fire regimes, demonstrating its promise for continent-scale, next-day wildfire spread prediction.}

\cite{deepa2024effective} proposes a contrastive learning framework for early forest fire prediction that combines a Contrastive Vision Transformer (CViT) with a Pool Former module. CViT functions as a powerful feature extractor via self-supervised contrastive learning and multi-head attention, while the Pool Former improves prediction efficiency by modeling spatial dependencies without heavy matrix operations. Though it does not employ traditional generative AI, the framework enhances feature robustness under variable environmental conditions through preprocessing and data augmentation strategies. \rev{The study utilized the publicly available Forest Fire Big Data dataset from Kaggle, comprising 1,832 images of forest scenes under diverse environmental and weather conditions, covering both fire and no-fire instances. In terms of quantitative performance, the CViT-Pool Former method achieved 92.8\% accuracy, 89.4\% precision, 91.4\% recall, and a 90.2\% F1-score, clearly outperforming Faster R-CNN (accuracy 89.0\%, F1 88.9\%) and ResNet (accuracy 87.4\%, F1 84.5\%). While the study did not report explicit training or inference times, it demonstrated that transformer-based contrastive learning significantly boosts predictive robustness and spatial discrimination compared to traditional deep CNN and RNN baselines.}

\cite{annane2024secured} introduces a real-time wildfire prediction system that integrates a CNN–Transformer hybrid model with a blockchain-based infrastructure for enhanced data security and traceability. The CNN extracts spatial features from drone imagery, while the transformer captures temporal patterns such as smoke visibility and fire direction. Although it does not use generative AI in the traditional sense, the system generates fire probability maps that simulate the spread and severity of wildfires, improving alert accuracy and resolution. It also incorporates a rule-based assistant for decision support and achieves 93.18\% prediction accuracy, outperforming CNN and ResNet-42 models, albeit with a slightly higher training time. \rev{The system was trained and tested on forest fire images from Algeria’s Béjaïa region, comprising 540 training images and 100 test images, each at a spatial resolution of 244×244 pixels. In quantitative terms, the CNN–Transformer model achieved an accuracy of 93.18\%, precision of 91\%, recall of 97\%, and an F1-score of 94\%, surpassing both standalone CNN and CNN ResNet-18 baselines. While it required a training time of approximately 30 minutes—slightly longer than the 20 minutes for CNN ResNet-18—this additional computational cost was offset by substantially higher predictive performance, particularly in modeling complex temporal features like fire evolution and direction. These results highlight the efficacy of integrating transformer-based temporal learning in enhancing the spatial-temporal resolution of wildfire risk maps.}

In contrast, \cite{li2024sim2real} presents a true generative AI approach through the development of Sim2Real-Fire, a large-scale, multi-modal dataset paired with the S2R-FireTr transformer model. S2R-FireTr forecasts and backtracks binary fire masks by learning from 1 million simulated wildfire sequences and generalizing to 1,000 real-world cases. By leveraging cross-attention across five aligned modalities—topography, vegetation, fuel types, weather, and satellite imagery—it generates physically plausible fire spread scenarios even under temporally incomplete conditions. 
\rev{Trained on 1 million multi-modal simulated wildfire scenarios with global coverage at 30 m spatial resolution, and evaluated on 1,000 annotated real-world wildfire cases, S2R-FireTr achieved an AUPRC of 72.9\%, F1-score of 69.6\%, and IoU of 56.4\% on real-world data. These results substantially outperformed traditional physical simulators—FARSITE (AUPRC 55.9\%), WFDS (61.2\%), and WRF-SFIRE (63.0\%)—as well as leading deep learning baselines like Rainformer and Earthformer. While exact training or inference times were not detailed, the approach demonstrated orders-of-magnitude efficiency gains by bypassing computationally intensive physics-based simulations, reinforcing its value for large-scale, real-time wildfire forecast and backtracking tasks.}

\rev{Together, these studies pave the way for broader adoption of advanced transformer-based and generative AI methods, which have likewise shown remarkable improvements over traditional simulation tools and conventional deep learning in wildfire forecasting. For example, \cite{li2024wildfire} reported that ASUFM, trained on 31,000 samples across North America, achieved superior Dice and PR-AUC metrics compared to U-Net and Swin U-Net, effectively modeling complex spatial fire dynamics at continental scale. Similarly, \cite{deepa2024effective} found that a contrastive vision transformer pipeline outperformed CNN and RNN baselines, reaching over 92\% accuracy and a 90\% F1-score on diverse forest imagery. \cite{annane2024secured} further demonstrated that integrating CNNs with transformers achieved 93\% accuracy and a 94\% F1-score on Algerian wildfire data, capturing temporal evolution beyond what CNNs alone could model. Finally, \cite{li2024sim2real} showed that a transformer trained on 1 million globally simulated wildfire scenarios attained AUPRC gains of 9–17 points over physics-based models like FARSITE and WFDS, enabling near real-time large-scale forecasting. Collectively, these results underscore the transformative advantages of GANs, VAEs, and transformers in enhancing the precision, scalability, and efficiency of wildfire spread prediction and behavior modeling.}

\subsection{Wildfire Detection and Monitoring }
Vision transformers are increasingly utilized in wildfire monitoring to generate high-resolution segmentation masks and predictive spatial representations of fire spread. These models excel in tasks requiring fine-grained spatial understanding and temporal consistency, such as detecting smoke plumes and segmenting active fire zones, due to the visual complexity of input data (e.g., RGB UAV imagery, thermal maps) and the operational need for accurate, real-time predictions to support early firefighting efforts. Our review identified three notable studies that incorporate transformer-based architectures to facilitate wildfire detection and monitoring.

\cite{falcao2023stacking} focuses on early detection of wildfire smoke plumes using a deep learning ensemble framework that integrates EfficientNetV2 (CNN), DeiT, and Swin TransformerV2 (both vision transformers). While generative models are not explicitly employed, the ensemble architecture—coupled with a neural network-based meta-classifier—demonstrates strong performance under challenging conditions like haze, fog, and low-contrast smoke. \rev{The study leverages transfer learning and data augmentation techniques to improve generalization, achieving an average accuracy of 96.46\% and an AUPRC of 95.14\%, improving upon the best individual base model by approximately 2.1\% in accuracy and 1.2\% in AUPRC. Precision remained comparably high to the strongest base learner (EfficientNetV2), while recall and F1-scores demonstrated consistent gains. Although the study did not benchmark against traditional machine learning or classical computer vision methods, it provided clear computational insights—training was performed via fine-tuning with transfer learning on a Google Colab platform using NVIDIA T4 GPUs, and inference time was measured at approximately 58 ms per batch of 16 images, compared to around 17 ms for individual models. This moderate increase in latency was justified by substantially enhanced robustness, allowing real-time deployment across hundreds of camera feeds operating at practical frame rates for wildfire monitoring.}

\cite{ghali2022deep} targets fire classification and segmentation using UAV imagery. Although it does not incorporate conventional generative AI, the framework produces fine-grained segmentation masks, effectively achieving a generative outcome. The authors develop an ensemble classifier with EfficientNet-B5 and DenseNet-201 for fire detection and employ three segmentation models—EfficientSeg, TransUNet, and TransFire—two of which are transformer-based. These models address complex challenges such as detecting small fires and delineating fire boundaries in noisy, cluttered backgrounds. \rev{The proposed transformer-based segmentation models demonstrated remarkable quantitative gains. TransUNet-R50-ViT achieved an accuracy and F1-score of 99.9\%, outperforming the CNN-based U-Net by nearly 1 percentage point, and effectively capturing fine fire boundaries even in cluttered backgrounds. TransFire also excelled with an F1-score of 99.82\%, confirming the advantage of vision transformers in extracting detailed fire regions. In terms of computational performance, TransUNet completed inference in 0.51 s per image, slightly higher than U-Net’s 0.29 s, while TransFire processed images at 1.0 s, reflecting a modest increase in latency for substantially improved segmentation fidelity. Although the study did not benchmark these models against traditional machine learning approaches, it clearly demonstrated the superior precision and robustness of transformer-based architectures in high-resolution wildfire segmentation using UAV imagery.}

\cite{ghali2021wildfire} explores the use of vision transformers—TransUNet and Medical Transformer (MedT)—to segment wildfire regions from RGB imagery for early detection and boundary delineation. These transformers perform generative tasks by producing binary segmentation masks that can be used to simulate wildfire spread scenarios. The models address core challenges such as fine boundary detection, long-range dependency modeling, and feature misclassification. Trained on the CorsicanFire dataset, TransUNet and MedT achieve F1-scores of 97.7\% and 96.0\% respectively, outperforming conventional architectures like U-Net, U2-Net, and EfficientSeg. Despite slightly longer inference times (1.2s for TransUNet and 2.72s for MedT), the trade-offs are justified by improvements in spatial granularity, generalization, and robustness under diverse environmental conditions. \rev{In terms of quantitative performance, TransUNet achieved an F1-score of 97.7\% with its hybrid ResNet50-ViT backbone, while MedT reached 96.0\%, each substantially exceeding the benchmarks set by U-Net (92.0\%), U2-Net (82.9\%), and EfficientSeg (94.3\%). Precision and recall were jointly reflected in these high F1-scores, underscoring the transformers’ capability to minimize both false positives and false negatives in fire boundary segmentation. Although the study did not report training time in detail, inference times were noted at 1.2 seconds for TransUNet and 2.72 seconds for MedT, representing a moderate computational overhead compared to simpler models, yet yielding superior segmentation fidelity and finer detection of small or diffuse fire regions.}

\rev{In terms of performance, transformer-based models have consistently outperformed traditional simulation tools and conventional deep learning approaches in both fire spread detection and boundary segmentation. For instance, \cite{falcao2023stacking} demonstrated that stacking vision transformers with CNNs boosted average accuracy to 96.46\% and AUPRC to 95.14\%, surpassing the best individual models by notable margins while maintaining practical inference times suitable for real-time monitoring. Similarly, \cite{ghali2022deep} and \cite{ghali2021wildfire} showed that transformer-based segmentation networks like TransUNet and TransFire achieved exceptional F1-scores up to 99.9\%, outperforming CNN-based baselines such as U-Net by 5–15 percentage points, with only modest increases in inference time. These results highlight the superior ability of transformers to capture fine spatial details and long-range dependencies, enhancing the precision, robustness, and spatial fidelity of wildfire detection and boundary modeling beyond what traditional or purely CNN-based methods can achieve.}
 
\subsection{Wildfire Risk Mapping}
Wildfire risk mapping plays a critical role in proactive fire management, enabling early warning systems, resource allocation, and strategic response planning across vulnerable landscapes. In this review, only one study using transformer-based method have been discovered.


 \rev{\cite{limber2024forecast} developed a residual transformer model to forecast wildfire potential across California by emulating the Wildland Fire Potential Index (WFPI), aiming to enhance short-term risk prediction and early warning for fire management. The model was trained to generate daily WFPI maps using Daymet meteorological data, MODIS-derived NDVI, and static fuel classifications from the Scott and Burgan fire models. As a generative forecasting emulator, it produces new spatial fire risk scenarios up to seven days ahead, addressing the high computational demands of physical models while ensuring rapid, spatially coherent forecasts. The residual connection leverages temporal autocorrelation in WFPI to boost short-term accuracy and stability. }
 
 \rev{As a performance summary, the model achieved spatial correlation coefficients of 0.85–0.98 across four weekly forecasts in July 2023, showing a slight tendency to underpredict extreme risk values. Precision remained high for one- and two-day forecasts, with modest degradation from days three to seven due to error propagation. Bayesian hyperparameter optimization and HPC resources enabled efficient tuning and training—requiring 42 hours on 48 GPUs for tuning and 24 hours on 24 GPUs for training—while full inference for four weeks of statewide daily forecasts was completed in just 6.5 minutes on a 128-core CPU. Although no direct comparisons were made to traditional or other ML models, the transformer substantially improved forecast speed and quality over existing USGS/USFS WFPI approaches, underscoring its promise for large-scale, data-driven wildfire risk prediction.}






\section{\rev{Discussion and Future Directions}}

\rev{Our targeted literature review identified and critically examined 11 studies that specifically explore the application of generative AI models to wildfire science. This relatively small number underscores that the integration of generative methods in this domain remains nascent, with substantial opportunities for further research and methodological innovation. Among the works reviewed, the predominant focus has been on leveraging GAN, VAE, and Transformer architectures to enhance bushfire prediction capabilities \cite{goodfellow2014generative, kingma2013auto, vaswani2017attention}. This clear research gap highlights a fertile area for advancing wildfire science through the development of more sophisticated, data-driven generative approaches that can simulate complex spatiotemporal fire dynamics and support robust decision-making. }

\rev{
Building on the preceding review of current generative AI applications in wildfire modeling and prediction, our analysis of 11 studies shows that generative AI models can achieve accuracies around 90\%, with some demonstrating computational performance exceeding traditional methods by several orders of magnitude. Combined with recent trends in LLM-powered agentic AI and distributed edge computing, these advantages open promising new research directions that remain largely unexplored. This section synthesizes key insights from the literature, outlines future avenues for investigation, and highlights major challenges along with potential strategies to address them.
}



\begin{figure}[H]
    \centering
    \includegraphics[width=1\textwidth]{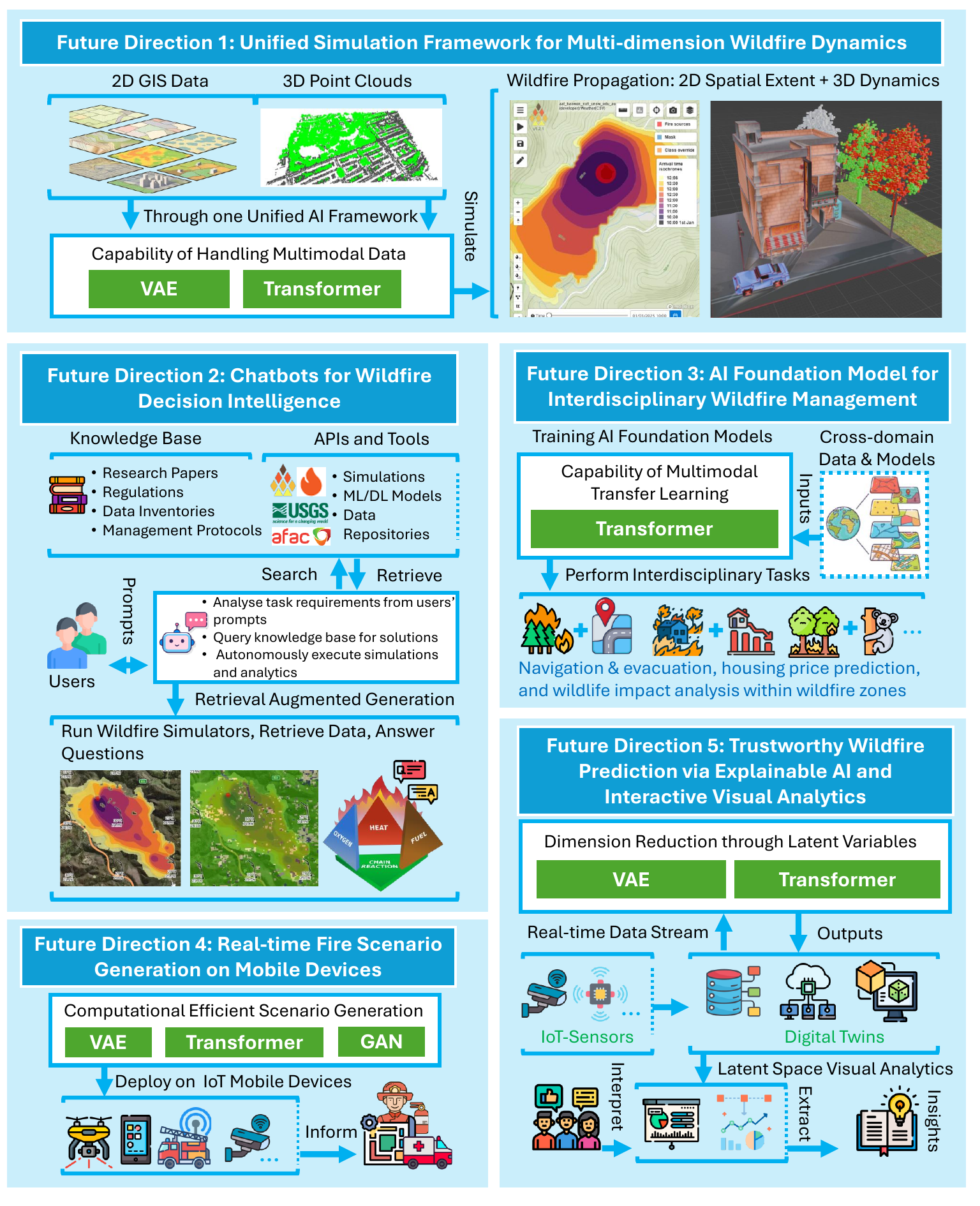}
    \caption{A summary of five future visions for applying generative AI to revolutionize wildfire prediction and management, ranging from the development of multimodal 2D and 3D wildfire modeling to the implementation of cognitive digital twins with explainable AI capabilities.}
    \label{fig:genAI_vision}
\end{figure}

\subsection{\rev{Future Research Directions} for Generative AI-powered Wildfire Applications}

 \rev{This section outlines a series of forward-looking research directions for how the existing generative AI-powered fire spread prediction applications can be further extended with more resaerch and development efforts to reshape wildfire prediction and management across scales and settings (as illustrated in Figure \ref{fig:genAI_vision}).}

\subsubsection{\rev{Unified Simulation Framework for 2D and 3D Wildfire Dynamic}}

Previous studies have highlighted the importance of using multimodal data to improve bushfire prediction, while also noting the significant technical challenges involved in fusing heterogeneous datasets \cite{shadrin2024wildfire, zakari2025enhanced}. Traditional wildfire modeling has primarily relied on 2D spatial inputs—such as satellite imagery, temperature maps, and vegetation indices—even though fire behavior is inherently three-dimensional, shaped by terrain elevation, vertical fuel structures, atmospheric dynamics, and built environments. Conventional deep learning approaches typically handle 2D and 3D data in isolation, requiring separate models for different modalities, which limits integration and adaptability.

\rev{
Despite the experimental use of transformers and VAEs for fire spread prediction in the 11 reviewed studies, current research has not explored their potential as multimodal generative frameworks for unifying diverse data sources. From a data science standpoint, architectures like VAEs and Transformers are inherently well-suited to integrate heterogeneous inputs by learning joint representations across modalities. VAEs can encode geospatial, meteorological, and 3D structural information into shared latent spaces governed by probabilistic distributions, while Transformers leverage self-attention to dynamically capture cross-modal and long-range dependencies \cite{li2025multimodal, xu2023multimodal}. This capability removes the need for explicit feature concatenation or separate fusion pipelines, enhancing scalability and allowing models to holistically capture the complex interplay of factors that drive wildfire propagation. By jointly learning how dynamic weather patterns interact with terrain and fuel characteristics, such approaches could improve predictive accuracy and enable scenario-based forecasting under varying meteorological conditions—critical for informed emergency response planning and efficient resource allocation. These considerations highlight a promising research agenda that leverages the unique strengths of multimodal generative AI to advance holistic wildfire simulation.
}

\subsubsection{\rev{Chatbots for Wildfire Decision Intelligence}}

\rev{Recent trends across various scientific disciplines, including transportation and climate science, increasingly leverage the integration of Large Language Models (LLMs) with Retrieval-Augmented Generation (RAG) to develop advanced conversational AI systems \cite{xu2024genai, xie2025wildfiregpt}. These multi-agent frameworks enable interactive, multi-turn dialogue that dynamically retrieves domain-specific data and generates context-aware, user-adaptive insights. Despite this growing popularity in other fields, such paradigms have yet to be explored for supporting wildfire decision intelligence.}

\rev{This represents a compelling research direction. Future work could explore how conversational AI assistants—built on fine-tuned LLMs, RAG mechanisms, and an agentic AI paradigm—might transform wildfire prediction and management \cite{xu2024genai, xu2025towards}. In such a framework, specialized agents could handle sub-tasks such as fire spread simulation, atmospheric analysis, or evacuation planning, all coordinated by a dialogue manager to ensure coherent, user-centered interactions. RAG would ground the assistant’s responses in real-time wildfire data and scientific knowledge, drawing on sources like localized fire weather indices and historical ignition patterns.}

\rev{By enabling natural language interaction, these AI assistants could empower urban planners, emergency responders, and community members to access relevant data, simulation outputs, and optimized decision solutions in an accessible format. This contrasts sharply with traditional systems that often rely on manual data processing and technical interfaces dense with domain-specific jargon. As an adaptive platform that learns continuously from expert feedback and evolving data streams, a wildfire-focused conversational AI system could facilitate more proactive, transparent, and inclusive approaches to fire risk assessment and emergency response. Advancing this vision represents an important future research agenda, leveraging the strengths of emerging LLM-RAG architectures to address longstanding challenges in wildfire decision support.}


\subsubsection{AI Foundation Model for
Interdisciplinary Wildfire Management}

\rev{
The emergence of AI foundation models—large-scale, pre-trained systems typically built on transformer architectures—has become an increasingly popular trend across numerous scientific domains, including remote sensing, urban logistics optimization, and GIS \cite{li2024empowering, myers2024foundation}. These models serve as general-purpose backbones trained on diverse datasets to support a wide range of downstream tasks, enabling cross-domain knowledge transfer, zero-shot inference, and rapid fine-tuning. Despite their growing success in other disciplines, foundation models remain largely unexplored in the context of wildfire science.}

\rev{
This gap represents a promising direction for future research. A foundation model for wildfire management could leverage generative AI architectures—such as LLMs, multimodal transformers, or diffusion networks—to integrate heterogeneous datasets across spatial, temporal, and semantic dimensions. Trained on inputs like satellite imagery, LiDAR scans, weather time series, terrain maps, fuel assessments, and incident reports, such a model could generate predictive outputs and scenario-based simulations under varying conditions, supporting applications ranging from fire spread forecasting to resource deployment planning.}

\rev{
Unlike traditional task-specific deep learning pipelines that often rely on separate models for different data modalities—such as CNNs for 2D raster GIS data—an AI foundation model would offer a unified, adaptive, and scalable framework. By inherently capturing cross-modal relationships, such a model could robustly generalize across diverse geographic regions, fire regimes, and management objectives. Integrating multimodal data within a single generative architecture would enable simultaneous analysis of evolving 2D and 3D spatial features, detection of smoke and vulnerable infrastructure from CCTV feeds, assessment of wildfire impacts on wildlife, and even prediction of housing price shifts in fire-prone areas, alongside automated synthesis of textual reports and advisories. Advancing this line of research represents a significant opportunity to enhance the deployability, scalability, and maintainability of AI-powered wildfire decision support, establishing a transformative paradigm for interdisciplinary wildfire management.
}

\subsubsection{\rev{Real-time Fire Scenario Generation on Mobile Devices}}
\rev{
Generative AI models such as VAEs and Transformers have shown considerable promise as computationally efficient surrogates for simulating wildfire spread. Unlike physics-based models that rely on intensive numerical solvers, these architectures can emulate spatiotemporal fire dynamics from historical or synthetic data, producing high-resolution wildfire scenarios with minimal computational overhead. Once trained, they offer near-instant inference, making them attractive for time-critical applications such as emergency response, evacuation planning, and real-time firefighting operations, and potentially deployable on mobile devices with limited computing resources.}

\rev{
However, despite the reviewed 11 studies demonstrating the computational efficiency and superior inference speeds of generative AI models, this research remains largely at an experimental stage. None of these studies have explored the bright potential of deploying such lightweight generative models directly on mobile platforms and edge devices. This represents an important gap and a promising direction for future research.
}

\rev{
Edge and mobile deployment could distribute computational and data analytics workloads across devices such as smartphones, UAVs, sensor controllers, and CCTV camera systems, reducing reliance on remote infrastructure and mitigating communication delays—especially critical in disaster-affected or connectivity-limited regions. Prior work has shown that convolutional and transformer-based models can be optimized for low-power hardware, such as NVIDIA Jetson Nano or Intel Movidius, enabling onboard inference without cloud dependence \cite{hu2024edge, mahdi2022edge}. Lightweight ViTs-based fire segmentation models are increasingly compatible with embedded systems, supporting high-frequency updates and real-time forecasting on resource-constrained devices \cite{spiller2022hyperspectral, lee2024vit}. 
}

\rev{
Future work could leverage these developments to embed generative wildfire models within mobile edge computing frameworks, enabling on-the-fly scenario generation using the most current field-collected data—such as wind speed, terrain, vegetation, and humidity. This would empower frontline responders with localized fire trajectory predictions, dynamic risk maps, and adaptive containment strategies, all processed directly on-site. Advancing this research agenda could establish a scalable, resilient, and connectivity-independent paradigm for delivering real-time wildfire intelligence.
}

\subsubsection{Trustworthy Wildfire Prediction via Explainable AI and Interactive Visual Analytics}
 
Recent advances in generative AI have opened new avenues for developing explainable and interactive wildfire prediction systems that go beyond traditional black-box models. Many state-of-the-art generative architectures, such as VAEs and Transformers, rely on latent spaces to encode complex environmental data into compressed representations. These latent variables offer a powerful mechanism for uncovering how models process input features and arrive at predictions or classifications \cite{xu2024explainable, liu2019latent}. Notably, although the 11 reviewed studies employed VAE- and transformer-based approaches that inherently learn latent representations and embeddings, none explored explainable AI techniques that visualize or interpret these latent spaces to clarify the models’ decision-making processes.

This gap highlights a compelling direction for future research. By developing visual analytics interfaces that directly examine latent spaces, stakeholders—including emergency managers, urban planners, and community members—could better understand how inputs such as fuel conditions, topography, and weather patterns collectively influence predicted fire spread. Interactive exploration of these low-dimensional representations would facilitate detection of meaningful patterns, anomalies, or risk clusters, enabling non-technical users to interrogate AI-driven forecasts and fostering greater transparency and trust. Such approaches stand in contrast to existing wildfire modeling tools that often produce opaque outputs requiring specialized interpretation.

\rev{
While these explainable AI systems could be embedded within broader platforms—such as urban digital twins or smart city dashboards—the core research agenda centers on advancing methodologies for visualizing, interpreting, and interacting with the internal decision processes of generative AI wildfire models. Integrating real-time data streams from IoT sensors, UAVs, or crowdsourcing platforms, these systems could continuously update their analyses and present the latest insights in an intuitive, user-centered format. Pursuing this line of research promises to significantly enhance the interpretability, reliability, and adoption of AI-based wildfire prediction tools, paving the way for more transparent and participatory fire risk management.}

 \begin{table}[htbp]
\centering
\caption{Challenges and potential solutions associated with applying generative AI models in wildfire prediction.}
\label{tab:challenges_solution}
\begin{tabular}{|p{3cm}|p{4.7cm}|p{4.7cm}|}
\hline
\textbf{Challenge} & \textbf{Description} & \textbf{Potential Solution} \\
\hline
\rev{Stochastic Nature of Wildfire Behavior} & \rev{Wildfire spread exhibits inherent unpredictability due to sensitivity to initial conditions, microclimate fluctuations, turbulent convection, and ember-driven spotting, which no model can fully eliminate.} & \rev{Employ ensemble simulations, scenario-based planning, and robust uncertainty quantification to capture probabilistic ranges of outcomes, supporting risk-informed decision-making rather than exact forecasts.} \\
\hline
Computational & Training and deploying large-scale generative AI models for wildfire prediction is resource-intensive due to high-resolution data demands, heavy memory usage, and slow inference. & Techniques such as quantization, low-bit training, distillation, and latent-space modeling can improve efficiency, though often at the cost of accuracy or interpretability. \\
\hline
Evaluation & Evaluating generative AI outputs is difficult due to the lack of standardized, interpretable, and domain-specific metrics for assessing the realism and utility of synthetic spatiotemporal fire scenarios. & Solutions include developing domain-specific and hybrid metrics, incorporating human-in-the-loop evaluation, uncertainty quantification, and physics-informed priors. \\
\hline
Energy and Environmental & Generative AI model development has a significant environmental impact due to high energy consumption and carbon emissions from large-scale training. & Mitigation strategies include model distillation, sparsity optimization, transfer learning, and adopting energy-efficient "green AI" practices \cite{salehi2023data, barbierato2024toward}. \\
\hline

\end{tabular}
\label{tab:ml_dl_genai_comparison}
\end{table}

\subsection{Challenges and Potential Solutions}
Despite the promising potential of generative AI in wildfire prediction, several critical challenges must be addressed to ensure its reliability, scalability, and scientific validity in real-world fire management applications. Table \ref{tab:challenges_solution} summarizes these key challenges along with potential solutions, which are further elaborated in the following subsections.

\subsubsection{\rev{Stochasticity Challenges in Fire Prediction}}

\rev{
While advances in computational modeling and data-driven learning have markedly improved our ability to simulate wildfire spread, it is critical to acknowledge the inherent stochasticity and irreducible unpredictability that characterize wildfire behavior \cite{sullivan2009wildland, morvan2011physical}. Unlike controlled physical systems, wildfires evolve within highly dynamic and often chaotic natural environments, where small variations in initial conditions—such as localized wind gusts, abrupt changes in humidity, or micro-scale fuel heterogeneity—can lead to dramatically different outcomes over time \cite{tedim2018defining, hilton2015effects}. This sensitivity to initial and boundary conditions underscores a fundamental limitation: even the most sophisticated models, whether physics-based or generative, cannot entirely eliminate uncertainty in predicting fire evolution.}

\rev{
Moreover, wildfires are influenced by a multitude of coupled processes across scales, including turbulent convection, ember lofting and spotting, and rapid transitions in combustion regimes, which are intrinsically probabilistic and only partially observable \cite{sullivan2009wildland}. For example, fine-scale wind vortices can transport embers far beyond modeled fire perimeters, triggering new ignitions in a manner that defies deterministic forecasting. Similarly, fuel moisture content can fluctuate over short distances due to microclimatic effects, introducing additional variability that is difficult to capture with static or coarse-grained input data \cite{hilton2015effects}.
}

\rev{
Consequently, while generative AI offers powerful new avenues for representing complex spread dynamics and for sampling from learned distributions that mirror observed variability, it does not circumvent the fundamental unpredictability embedded in wildfire phenomena. Instead, these models should be viewed as tools to better characterize the probabilistic ranges of likely outcomes and to improve computational efficiency, rather than to produce more precise point forecasts. This perspective emphasizes the need for ensemble simulations, scenario-based planning, and robust uncertainty quantification as integral components of any wildfire prediction framework \cite{rochoux2018front, hilton2015effects}. By explicitly accounting for the stochastic nature of fire behavior, researchers and practitioners can make more informed decisions that appropriately balance forecast precision with the intrinsic volatility of wildfire systems.
}

\subsubsection{Computational Challenges}
Training large-scale generative AI models—particularly diffusion models, transformers, and other autoregressive architectures—presents substantial computational challenges. These models typically require high-resolution spatial-temporal data spanning vast geographic regions and extended fire seasons, leading to massive, memory-intensive datasets. As emphasized by Manduchi et al.~\cite{manduchi2024challenges}, training such models at scale demands access to powerful GPUs or TPUs, careful memory management, and optimization strategies to maintain training stability across distributed systems.

Moreover, inference with diffusion models often involves iterative denoising steps, while transformer-based language models follow a sequential token generation process, both of which result in high latency. This is particularly problematic in real-time or near-real-time wildfire forecasting applications where rapid decision-making is essential. Even recent improvements like FlashAttention and latent-space diffusion training~\cite{manduchi2024challenges, bandi2023power} only partially mitigate these bottlenecks. Additionally, robust training requires extensive data augmentation, domain adaptation, and fine-tuning, which further increases the computational burden~\cite{bandi2023power}.

To improve computational efficiency, emerging strategies such as model quantization, low-bit training, distillation, and lossy latent-space modeling have shown promise~\cite{manduchi2024challenges}. However, these approaches often come with trade-offs in terms of accuracy, robustness, or interpretability—making them challenging to adopt in high-stakes domains such as environmental hazard forecasting and disaster response.

\subsubsection{Evaluation Challenges}
Evaluating the outputs of generative AI models in wildfire prediction poses significant methodological challenges, particularly due to the lack of standardized, interpretable, and domain-specific evaluation frameworks. Unlike classification or regression tasks, where performance can be quantified using well-established metrics, the assessment of synthetic wildfire scenarios—especially those generated under creative or hypothetical conditions—remains highly subjective and context-dependent \cite{bandi2023power, manduchi2024challenges}. Widely used metrics such as Fréchet Inception Distance (FID) or Inception Score (IS), originally developed for image synthesis, are ill-suited for evaluating spatiotemporal fidelity, physical realism, or consistency with known fire behavior dynamics in environmental simulations \cite{bandi2023power, manduchi2024challenges}.

Recent studies emphasize the urgent need for developing ``domain-specific evaluation metrics'' tailored to complex generative tasks. These include assessing spatial spread accuracy, temporal progression, and physical plausibility relative to empirical wildfire data and simulation-based fire behavior models \cite{fui2023generative}. Additionally, generative AI models frequently suffer from pathologies such as ``mode collapse, sample hallucination, and memorization'', which compromise their generalizability and reliability when trained on sparse or biased fire datasets \cite{manduchi2024challenges}. These issues are particularly critical in ``high-stakes applications'' such as real-time emergency response or predictive decision support, where misleading outputs can result in dangerous operational consequences.

The evaluation challenge is further exacerbated by the absence of ground truth for rare or extreme wildfire events, which limits model validation using traditional benchmarks. As \cite{manduchi2024challenges} and \cite{sun2024generative} argue, current benchmarks fail to capture domain-specific constraints and interpretability, especially in dynamic, real-world environments. To address these shortcomings, researchers have advocated for the integration of ``human-in-the-loop evaluation'', ``uncertainty quantification'', and ``physics-informed priors'' into both model training and assessment pipelines \cite{yan2024promises, fui2023generative}. Furthermore, a growing body of literature calls for ``hybrid metrics'' that combine statistical quality measures with expert-driven assessments and simulation-based validations, offering a more holistic evaluation of generative outputs \cite{bandi2023power, manduchi2024challenges, fui2023generative}.

\subsubsection{Energy and Environmental Challenges}
One significant but often under-discussed challenge in developing generative AI models for wildfire prediction lies in their substantial energy consumption and environmental footprint. Training foundation models—particularly those based on large-scale transformer architectures—requires extensive computational resources, often involving thousands of GPU hours and high-performance computing clusters \cite{zhou2024training}. This process results in considerable electricity usage, much of which is still powered by fossil fuels in many regions, leading to high levels of carbon emissions. As highlighted by \cite{myers2024foundation}, the environmental cost of training a single large language model can surpass the annual carbon footprint of several individuals. While these models offer valuable capabilities for simulating fire dynamics and generating proactive mitigation strategies, their development raises important concerns regarding sustainability and ecological responsibility, especially in the context of wildfire prediction, where climate change and ecosystem degradation are already pressing issues.

To mitigate these impacts, it is essential to explore strategies such as model distillation, sparsity optimization, transfer learning from pre-trained models, and leveraging green AI principles that emphasize energy-efficient training and inference \cite{salehi2023data, barbierato2024toward}. Integrating these strategies not only aligns the development of AI models with climate-conscious values but also ensures that the tools designed to protect the environment do not inadvertently contribute to its degradation.

\section{Conclusion}
\rev{This review systematically examined the emerging role of generative AI models—including VAEs, GANs, Transformers, and diffusion architectures—in advancing bushfire prediction, monitoring, and risk mapping. While only 11 studies were identified that specifically explore these approaches for wildfire science, their collective findings demonstrate that generative AI can achieve prediction accuracies near 90\% and offer computational efficiencies far surpassing traditional methods. Despite these advances, the use of generative AI in this domain remains nascent compared to its widespread adoption in other scientific and engineering fields.}

Building on these observations, we proposed several promising research directions to shape the next generation of wildfire decision support systems. These include developing unified multimodal frameworks that seamlessly integrate 2D and 3D data, designing conversational agentic AI systems to deliver interactive, real-time wildfire intelligence, training interdisciplinary AI foundation models, enabling edge-based scenario generation on mobile and IoT devices, and advancing explainable AI interfaces for improved transparency and trust.

Addressing these frontiers will require tackling notable challenges such as managing inherent wildfire stochasticity, reducing computational and energy footprints, and establishing rigorous, domain-specific evaluation standards. By harnessing the unique strengths of generative AI while embedding principles of explainability, sustainability, and interdisciplinary integration, future research can build robust, adaptive, and human-centered systems that significantly enhance wildfire prediction and emergency management capabilities.

\small
\bibliographystyle{elsarticle-harv}
\bibliography{src}

\end{document}